\documentclass[lettersize,journal]{IEEEtran}
\usepackage{amsmath,amsfonts}
\usepackage{algorithmic}
\usepackage{array}
\usepackage[caption=false,font=normalsize,labelfont=sf,textfont=sf]{subfig}
\usepackage{textcomp}
\usepackage{stfloats}
\usepackage{url}
\usepackage{verbatim}
\usepackage{graphicx}
\usepackage{pifont}
\newcommand{\cmark}{\ding{51}} 
\newcommand{\xmark}{\ding{55}} 
\usepackage{multirow}
\def\BibTeX{{\rm B\kern-.05em{\sc i\kern-.025em b}\kern-.08em
    T\kern-.1667em\lower.7ex\hbox{E}\kern-.125emX}}
\usepackage{balance}
\usepackage{cite}
\usepackage[backref]{hyperref}
\hypersetup{
    colorlinks=true,
    linkcolor=red,
    filecolor=blue,      
    urlcolor=blue,
    citecolor=blue,
}
\usepackage[table]{xcolor}
\definecolor{brightpink}{RGB}{255,20,147}
\begin{document}
\title{\huge
Focus on What Really Matters in Low-Altitude Governance:
A Management-Centric Multi-Modal Benchmark with Implicitly Coordinated Vision-Language Reasoning Framework
}

        \author{
        Hao~Chang,
        Zhihui~Wang,
        Lingxiang~Wu,
        Wei~An,
        Boyang~Li,
        Zaiping~Lin,
        Weidong~Sheng,
        Jinqiao~Wang
        \thanks{This work was supported by the National Natural Science Foundation of China under Grant 62501618.
        Corresponding author: Lingxiang~Wu and Wei~An (e-mail: lingxiang.wu@nlpr.ia.ac.cn, anwei@nudt.edu.cn).}
        \thanks{Hao Chang and Wang Zhihui contributed equally to this work.}
        \thanks{Hao Chang, Boyang~Li, Zaiping~Lin, Weidong~Sheng and Wei~An 
        are with the National University of Defense Technology, China.}
        \thanks{Zhihui Wang is with Zidong Taichu (Beijing) Technology Co., Ltd., China.}
        \thanks{Lingxiang Wu is with Zidong Taichu (Beijing) Technology Co., Ltd., China, 
        and also with the Foundation Model Research Center, 
        Institute of Automation, Chinese Academy of Sciences, China.}
        \thanks{Jinqiao Wang is with Zidong Taichu (Beijing) Technology Co., Ltd., China, 
        and also with the Foundation Model Research Center, Institute of Automation, Chinese Academy of Sciences, China, 
        also with the School of Artificial Intelligence, University of Chinese Academy of Sciences, China, 
        and Wuhan AI Research, China.}
        }


\maketitle

\begin{abstract}
Low-altitude vision systems are becoming a critical infrastructure for smart city governance.
However, existing object-centric perception paradigms and loosely coupled vision–language pipelines are still difficult to support management-oriented anomaly understanding required in real-world urban governance.
To bridge this gap, we introduce GovLA-10K, the first management-oriented multi-modal benchmark for low-altitude intelligence,
along with GovLA-Reasoner, a unified vision-language reasoning framework tailored for governance-aware aerial perception.
Unlike existing studies that aim to exhaustively annotate all visible objects, 
GovLA-10K is deliberately designed around functionally salient targets that directly correspond to practical management needs, 
and further provides actionable management suggestions grounded in these observations.
To effectively coordinate the fine-grained visual grounding with high-level contextual language reasoning, 
GovLA-Reasoner introduces an efficient Spatially-aware Grounding Adapter (SGA) that 
implicitly coordinates discriminative representation sharing between the visual detector and the large language model (LLM).
Different from existing adapters that primarily focus on global embedding alignment,
our SGA is specifically designed to compress and aggregate multi-stream grounding-aware representations, 
thereby preserving fine-grained spatial cues while enabling their effective integration into the language reasoning process.
Extensive experiments indicate that our GovLA-Reasoner effectively improves performance 
while avoiding the need of fine-tuning for any task-specific individual components.
We believe our work offers a new perspective and foundation for future studies on management-aware low-altitude vision–language systems.
\textcolor{brightpink}{The code and dataset will be publicly released after further organization.}
\end{abstract}


\section{Introduction} \label{sec1}
\subsection{Background}
\IEEEPARstart{L}{ow-altitude} vision systems, 
enabled by unmanned aerial vehicles (UAVs) and deep learning techniques, 
are increasingly deployed in recent years \cite{9615243,11457351,9417704,11251127}.
Compared with conventional ground-based perception systems, 
low-altitude platforms offer a unique vantage point with broader spatial coverage and real-time awareness, 
making them particularly valuable for large-scale urban management and emergency response \cite{wang2025uav,yuhong2025towards,wang2025technological,jiang2025review}.

Existing research in low-altitude predominantly targets general-purpose visual perception, 
such as image classification \cite{deng2025lightweight,al2025multi,zhang2025attention}, object detection \cite{xu2025real,alshehri2025integrated,zhou2025uav}, object tracking \cite{xu2025online,yao2025mm,liu2025multi}, and semantic segmentation \cite{chen2025lightweight,yang2025rshrnet,huang2025expanding}.
These studies aim to improve the completeness and accuracy of visual recognition, 
and have established the strong technical foundation for understanding low-altitude scenes \cite{ren2016faster,cai2018cascade,ge2021yolox,zhu2020deformable,zhang2022dino}.
To support this paradigm, most publicly available low-altitude datasets adopt a generic, detection-oriented annotation strategy 
that labels all visible interest objects in a scene \cite{hsieh2017drone,bozcan2020air,du2019visdrone,yu2020scale,shermeyer2021rareplanes,sun2022drone}.
Such benchmarks are highly suitable for full-scene coverage, enabling effective and comprehensive visual understanding.

Particularly, with the rapid emergence of large language models (LLMs) and large vision-language models (VLMs), 
the paradigm of low-altitude scene understanding is undergoing a significant transformation.
Unlike conventional task-specific models that rely on limited-scale annotated datasets, these foundation models are pretrained on massive and diverse multimodal corpora, 
endowing them with strong generalization and knowledge transfer capabilities.
As the result, many LLM/VLM-based systems \cite{dai2023instructblip,zhu2023minigpt,team2023gemini,team2024gemini,Qwen2.5-VL,LLaVA-OneVision-1.5,zhu2025internvl3,wang2025internvl3_5,Qwen3-VL} 
demonstrate impressive zero-shot performance in low-altitude scenarios, 
which can accurately recognize common aerial targets and provide semantically coherent interpretations without requiring task-specific optimization.

\subsection{Challenge}
However, the requirements of real-world low-altitude management fundamentally differ from exhaustive visual perception.
In practical urban governance, the target is not to recognize all objects indiscriminately, 
but to selectively identify abnormal, risky, or regulation-violating situations, 
for example, illegally parked vehicles, construction safety hazards, or sanitation-related risks.
This shift from comprehensive recognition to selective anomaly confirmation demands higher-level semantic interpretation and situational awareness, 
which go beyond the scope of traditional object-centric perception.
Therefore, the key challenge in real-world low-altitude management does not lie in further pushing the lower bound of identification accuracy, 
which has already been well addressed by existing visual detection models. 
Instead, the bottleneck arises from the lack of a management-oriented benchmark dataset that explicitly guide models which objects require attention and why they matter in urban governance.

Beyond this gap, another critical limitation in current low-altitude systems lies in the insufficient coordination between visual detectors and VLMs.
Specifically, since VLMs are unlikely to accurately recognize designated targets in every low-altitude urban scenario, 
deploying scenario-optimized visual detectors to compensate for this limitation is a practical solution. 
However, existing pipelines generally adopt a loosely coupled paradigm, 
in which the predicted bounding boxes and images are converted into structured, fine-grained prompts for VLMs, 
with the aim of achieving more precise object grounding and scene understanding.
However, this explicit coordination paradigm inevitably introduces non-negligible side effects.
The grounding information extracted by vision detectors must be reinterpreted, structured, and transmitted across heterogeneous stages, 
during which intermediate representations are prone to information loss and error accumulation.
As the result, even minor inaccuracies in visual localization or contextual abstraction at the grounding stage can significantly impact subsequent cross-modal interactions,
ultimately leading to substantial degradation in downstream language-based understanding and decision-making performance.
Therefore, introducing a more integrated and principled coordination mechanism between visual perception and language reasoning still needs more research.

\subsection{Solution}
To address these limitations, 
we propose a coordinated benchmark-framework approach that jointly redefines data annotation and model architecture to satisfy the practical demands of real-world low-altitude management.

On benchmark side, we introduce \textbf{GovLA-10K}, the first multimodal dataset tailored for management-oriented low-altitude scenarios.
In contrast to existing low-altitude datasets that primarily emphasize exhaustive object-level annotations for generic visual perception,
GovLA-10K is deliberately constructed from the perspective of urban governance,
focusing on functionally salient targets associated with abnormal, risky, and regulation-violating situations that truly require managerial attention.
Beyond selective visual annotations, GovLA-10K further provides aligned language-based scene descriptions and management suggestions grounded in visual evidence.
By shifting the supervision objective from full-scene recognition to selective, semantics-driven event understanding,
our GovLA-10K establishes a solid data foundation for low-altitude vision systems.

Beyond benchmark construction, we further propose \textbf{GovLA-Reasoner}, 
a unified vision–language reasoning framework that establishes a more integrated and principled coordination mechanism between visual perception and language reasoning, 
overcoming the limitations of loosely coupled detector–VLM pipelines.
Specifically,
instead of explicitly converting localized predictions into structured prompts, 
GovLA-Reasoner introduces an efficient Spatially-aware Grounding Adapter (SGA) that 
effectively aggregates discriminative feature representations from the visual detector
and further directly feeds them into LLM for high-level contextual reasoning. 
This implicit feature-space coordination enables end-to-end interaction between vision and language, 
avoiding fragile intermediate representations and reducing information loss.
Peculiarly, unlike existing commonly used adapters such as Q-Former \cite{li2023blip} or LLaVA-style Projector \cite{liu2023visual} that primarily focus on global embedding alignment,
these designs are not well suited for handling the multi-stream localization features produced by detectors.
In such representations, complementary spatial cues are distributed across heterogeneous feature sources.
Our SGA is specifically designed for localization-aware feature integration. 
It compresses and coordinates multi-stream grounding features while preserving spatially discriminative cues, 
enabling structured detection representations to be seamlessly transformed into language-compatible embeddings.

Notably, our GovLA-Reasoner is efficient and deployment-friendly. 
It delivers effective performance gains without fine-tuning either the visual detector or the language model.
Besides, compared with existing adapters, the adopted SGA achieves a better trade-off between performance and computational cost.
These two advantages make it well suited for resource-constrained low-altitude intelligence applications.
In summary, our work fills a critical gap in low-altitude vision research, with the following key contributions:
\begin{itemize}
\item \textbf{Benchmark Contribution:} We introduce GovLA-10K, the first management-oriented multimodal benchmark for low-altitude vision systems, 
shifting the research focus from exhaustive object recognition to selective and actionable understanding of abnormal and risky events.

\item \textbf{Framework Contribution:} We propose GovLA-Reasoner, 
a unified vision–language reasoning framework that harmonizes visual perception and language reasoning via implicit feature-space coordination, 
effectively mitigating error accumulation and information loss inherent in existing loosely coupled pipelines.

\item \textbf{Practical Significance:} By jointly leveraging GovLA-10K and GovLA-Reasoner, 
we present an efficient and deployment-friendly solution for low-altitude intelligence. 
Our framework achieves effective performance gains while better controlling the computational resource cost,
establishing a new paradigm for governance-oriented aerial vision systems.
\end{itemize}

\begin{figure*}[!t]
    \centering
    \includegraphics[width=\linewidth]{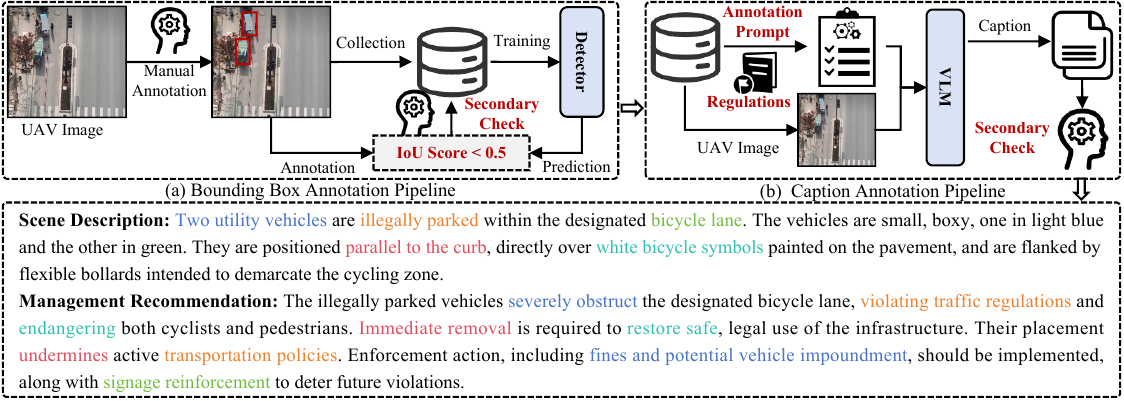}
    \caption{
Annotation pipeline of GovLA-10K. 
Our GovLA-10K adopts a two-stage semi-automatic annotation pipeline. 
(a) Bounding box annotation stage: interest regions are manually annotated and cross-verified using a powerful detection model to ensure annotation accuracy. 
(b) Caption annotation stage: verified box annotations and original images are converted into structured, fine-grained prompts and fed into the VLM to generate contextual captions. 
Peculiarly, all generated captions are also further reviewed by human annotators to maintain high annotation quality. 
}
    \label{data1}
\end{figure*}

\section{Benchmark: GovLA-10K}
As mentioned in Section \ref{sec1}, one major limitation of existing low-altitude system research is 
the lack of management-oriented benchmark that can explicitly guide models to identify which objects deserve attention 
and why they are important in the context of urban governance. 
To address this gap, we construct GovLA-10K dataset.
The detailed information of the dataset is presented as follows.

\subsection{Data Collection and Annotation}
\subsubsection{Data Collection.}
The construction of GovLA-10K is guided by the practical demands of real-world low-altitude management, 
where UAVs act as the primary sensing platform. 
We therefore focus on collecting low-altitude UAV images that cover diverse urban living environments and governance-relevant scenarios, 
including complex urban layouts, construction activities, traffic conditions, and sanitation-related phenomena.

To ensure both diversity and realism, GovLA-10K is built upon the data collected from two complementary sources:
(1) large-scale crawling of publicly available low-altitude aerial images from online platforms and UAV communities, and 
(2) in-house UAV flights along predefined routes over representative urban areas such as residential communities, commercial districts, construction zones, and traffic-intensive regions. 
In total, we initially gather about 16,000 raw UAV images.
All images are uniformly cropped into $512 \times 512$ patches and further filtered to remove low-quality, redundant, or weakly relevant samples. 
After screening, 10,572 high-quality images are retained. 

Based on extensive investigation of urban management practices and the visual distribution of collected data, 
we identify nine functionally salient categories that directly correspond to common low-altitude governance concerns, 
including: illegally parked vehicle, construction debris, construction fencing, brick pile, aggregate pile, construction worker, ground litter, overflowing trash bin, and scaffolding. 
These categories deliberately emphasize abnormal, risky, or regulation-related targets, rather than exhaustive object inventories.
\subsubsection{Data Annotation.}
To achieve both efficiency and precision, we design a two-stage semi-automatic annotation pipeline that integrates powerful visual detectors with VLMs.
As illustrated in Figure \ref{data1}, in the first stage, 
several domain experts manually annotate the horizontal bounding boxes (BBoxes) and category labels for the selected targets. 
After the initial annotation, we train a strong grounding-oriented visual detector based on MM-GroundingDINO \cite{zhao2024open}. 
This detector is then applied to the entire dataset, and its predictions are compared with the original human annotations by computing Intersection-over-Union (IoU) scores. 
All instances with IoU lower than 0.5 are automatically collected and sent back for secondary manual inspection and correction. 
This cross-verification mechanism enables rapid large-scale checking while effectively preventing systematic annotation bias and localization errors.

In the second stage, we aim to provide fine-grained semantic descriptions and management-oriented interpretations for each annotated scene.
Since the BBox annotations are manually verified and thus do not suffer from the significant localization noise,
the explicit prompting paradigm is feasible at this stage.
Specifically, the verified BBox annotations and original images are converted into structured, 
fine-grained prompts, which are jointly fed into the state-of-the-art open-source VLM (Qwen3VL-235B-A22B \cite{Qwen3-VL}) to generate contextual captions. 
In addition, the prompt template is carefully constrained to suppress stylistic dominance and maintain consistent, visually grounded descriptions.
Each caption includes not only an objective scene description grounded in visual evidence, but also management recommendations aligned with urban governance logic.
Moreover, we further incorporate relevant regulations into the prompts to improve the reliability and governance consistency of the generated recommendations and mitigate potential hallucinations.
All generated captions are further reviewed by expert annotators and cross-checked against visualized BBoxes to ensure semantic correctness, completeness, and practical relevance.


\begin{figure*}[!t]
    \centering
    \includegraphics[width=\linewidth]{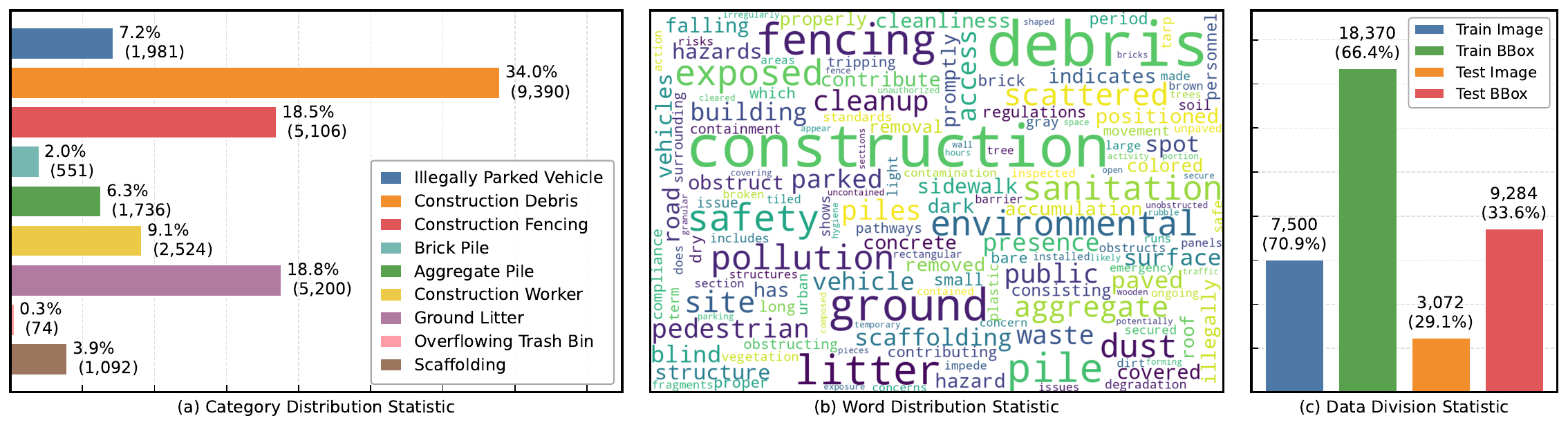}
    \caption{
    Basic statistics of GovLA-10K. 
    (a) Category-wise instance counts and proportions. 
    (b) High-frequency word distribution in data captions. 
    (c) Numbers of images and bounding boxes in training and test sets.
    }
    \label{data2}
\end{figure*}

\subsection{Data Statistic}
\subsubsection{BBox Statistic.}
Figure~\ref{data2} (a), (c), and Table~\ref{tab_data2} summarize the instance statistics of the nine functional categories in GovLA-10K.
Construction Debris (34.0\%), Ground Litter (18.8\%), and Construction Fencing (18.5\%) dominate the dataset, reflecting their prevalence and governance importance in urban low-altitude scenarios, 
whereas Overflowing Trash Bin (0.3\%), Brick Pile (2.0\%), and Scaffolding (3.9\%) are relatively rare, consistent with real-world distributions.
The remaining categories occupy moderate proportions, covering common urban management concerns.
We split the dataset into training and testing sets at an approximate 7:3 ratio (7,500 / 3,072 images) using stratified sampling to ensure representative coverage of all categories, especially rare ones.

\subsubsection{Caption Statistic.} 
\begin{figure}[t]
    \centering
    \includegraphics[width=\linewidth]{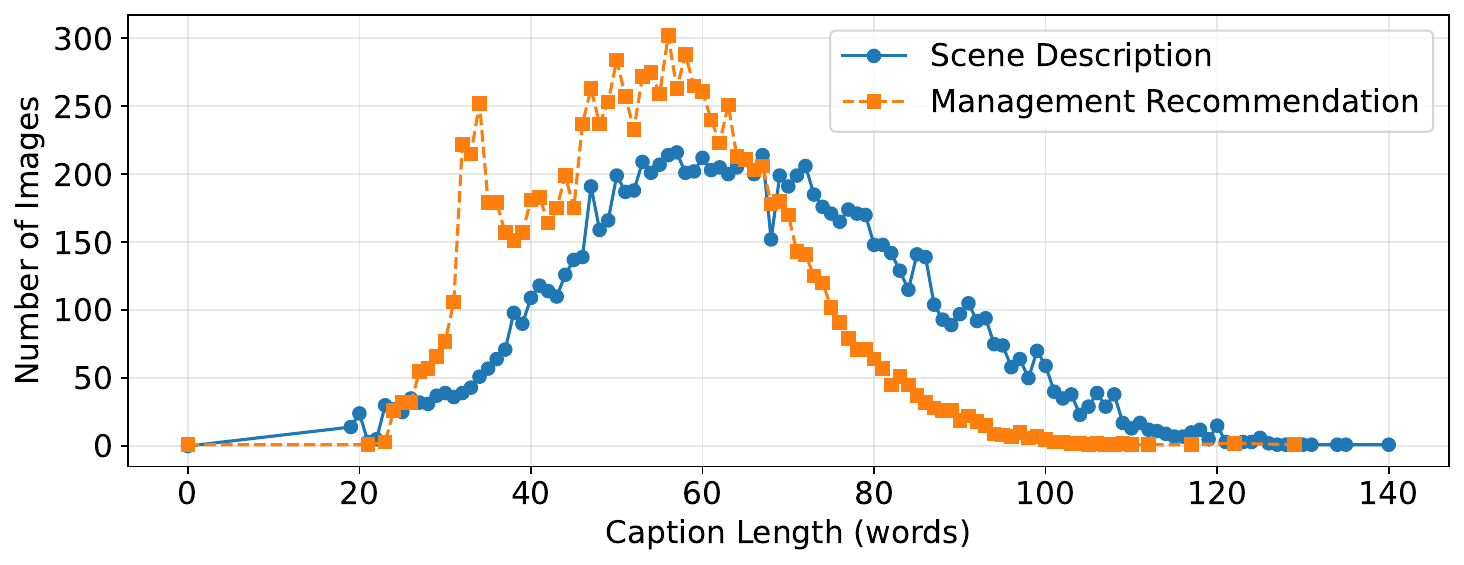}
    \caption{Distribution of image numbers across different caption lengths.
    }
    \label{data3}
\end{figure}

Figure~\ref{data2} (b) depicts the distribution of high-frequency keywords in generated captions. 
The results indicate that captions are closely aligned with the annotated target categories and effectively incorporate management-relevant concepts critical for low-altitude urban governance,
such as safety, public, and sanitation. 
Figure~\ref{data3} further illustrates the distribution of image counts across different caption lengths, 
which are predominantly concentrated around 60 words. 
This range allows comprehensive coverage of key targets and scene details while avoiding excessive verbosity, 
ensuring the overall quality and utility of the captions. 
These statistics once again provide clear evidence of the effectiveness of our caption annotation pipeline.

\begin{table}[t]
\centering
\caption{Detailed category information statistic of GovLA-10K.}
\setlength{\tabcolsep}{8pt}
\renewcommand{\arraystretch}{1.2}
\resizebox{\linewidth}{!}{
\begin{tabular}{l|c|cc}
\hline \hline
Category &Proportion  & Training Set & Testing Set \\
\hline 
Illegally Parked Vehicle &7.2\% & 1,565 & 416 \\
Construction Debris  &34.0\%      & 5,794 & 3,596 \\
Construction Fencing  &18.5\%     & 3,284 & 1,822 \\
Brick Pile    &2.0\%             & 365   & 186 \\
Aggregate Pile   &6.3\%          & 1,082 & 654 \\
Construction Worker &9.1\%    & 1,619 & 905 \\
Ground Litter &18.8\%              & 3,908 & 1,292 \\
Overflowing Trash Bin &0.3\%     & 53    & 21 \\
Scaffolding   &3.9\%            & 700   & 392 \\
\hline \hline
\end{tabular}
}
\label{tab_data2}
\end{table}

\subsection{Scenario Analysis} 
\begin{figure*}[!t]
    \centering
    \includegraphics[width=\linewidth]{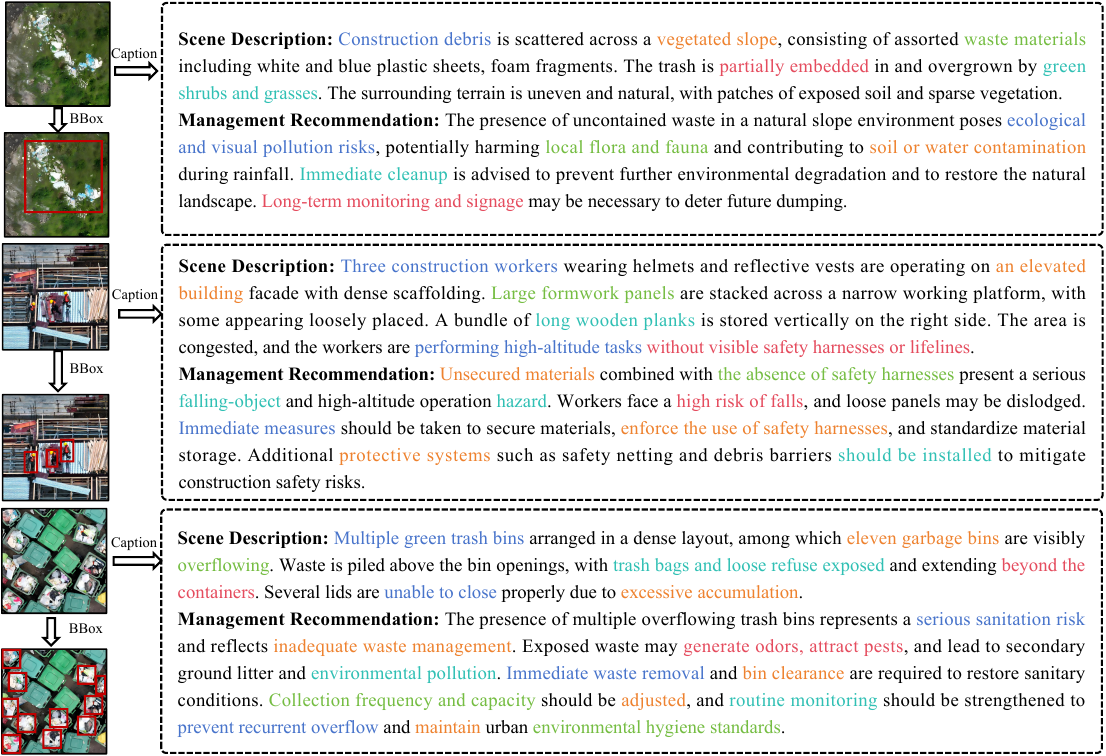}
\caption{Some representative scenarios of GovLA-10K.
The key contents are highlighted in color for clarity.
 }
    \label{demo2}
\end{figure*}

Figure \ref{demo2} displays some representative examples in our GovLA-10K dataset. 
As illustrated, GovLA-10K consistently focuses on management-oriented targets in low-altitude urban scenes, 
rather than indiscriminately annotating all visible objects. 
This design choice highlights the practical relevance of the dataset for real-world urban governance, 
where only specific objects and areas are truly critical for management and decision-making.
Moreover, the fine-grained captions provided in GovLA-10K go beyond objective visual descriptions of the scenes. 
They further incorporate management-related interpretations and corresponding recommendations, 
enabling a closer alignment between visual perception and urban governance requirements. Such governance-aware annotations make GovLA-10K not only a perception benchmark, but also a foundation for high-level reasoning and management-oriented understanding.
These qualitative results intuitively demonstrate the effectiveness of our designed two-stage semi-automatic annotation pipeline, 
which is able to reliably identify management-relevant targets and generate structured, governance-aware descriptions at scale.

\section{Framework: GovLA-Reasoner}
\begin{figure*}[!t]
    \centering
    \includegraphics[width=\linewidth]{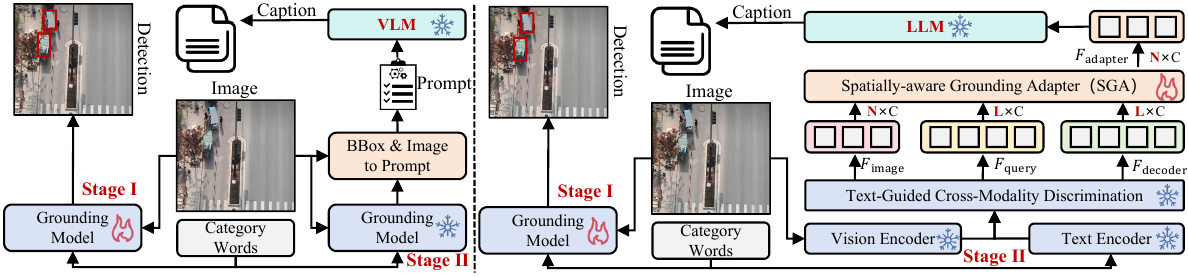}
    \caption{
    Low-altitude framework overview of the existing pipeline (left) and our proposed GovLA-Reasoner (right). 
    To address the information loss and error accumulation problems caused by the explicit prompt construction, 
    GovLA-Reasoner adopts an implicit feature-space coordination paradigm and further introduces SGA, 
    which adaptively compresses and aggregates the multi-stream discriminative features of grounding model to directly provide the robust feature representation for LLM.
    Note: snowflakes indicate frozen parameters, whereas flames indicate active (trainable) parameters.
    }
    \label{net}
\end{figure*}

Figure \ref{net} illustrates the detail differences between the existing loosely coupled reasoning pipeline and our proposed GovLA-Reasoner. 
In existing paradigms, 
the detection model is first optimized for specific scenarios to ensure the precise target grounding. 
Then, the detection outputs are combined with the original images to construct fine-grained, structured prompts, 
which are subsequently fed into VLMs to produce detailed descriptions. 
Although effective, this explicit pipeline inevitably leads to information loss and error accumulation.
To address this limitation, our GovLA-Reasoner introduces an efficient Spatially-aware Grounding Adapter (SGA),
which can achieve the more coherent end-to-end vision-language coordination.
Below is the detailed design of GovLA-Reasoner.

\subsection{Stage \MakeUppercase{\romannumeral 1}: Visual Grounding}
As mentioned in Section \ref{sec1}, 
although VLMs exhibit strong generalization capabilities, 
they are unlikely to precisely recognize potential targets in every real-world low-altitude scenario. 
This limitation directly affects the reliability of subsequent language understanding and scene analysis. 
Therefore, the primary goal in Stage I is to optimize the vision detector for specific low-altitude scenarios, 
providing a stable visual foundation for caption generation.

Considering that existing vision detectors have already achieved strong performance,
we do not focus on redesigning detector architectures, but on selecting the most suitable model for the current low-altitude management scenario. 
In this process, our dataset, GovLA-10K, differs significantly from conventional low-altitude datasets: 
it is designed around practical urban governance needs, focusing only on abnormal or risky targets instead of exhaustive annotations. 
This ensures that the grounding model is well-adapted to the task.
Compared with conventional pure-vision detectors that only take images as input, grounding models offer two key advantages.
First, they can take textual prompts as input, 
naturally enabling selective and open-set recognition, 
which aligns with both the design of GovLA-10K and the practical requirements of urban management. 
Second, the generated discriminative features are naturally enhanced through vision-language cross-modal interactions, 
providing a solid foundation for implicit feature alignment and collaborative reasoning with LLMs.
\subsection{Stage \MakeUppercase{\romannumeral 2}: Caption Reasoning}
Building upon the precise visual grounding obtained in Stage I, 
Stage \MakeUppercase{\romannumeral 2} aims to efficiently receive and exploit grounding-aware visual representations to generate accurate and reliable captions.
Instead of following the loosely coupled pipeline that converts detection outputs into structured prompts for VLMs, 
GovLA-Reasoner introduces an efficient adapter to directly transform grounding features into LLM-compatible representations.

Specifically, existing explicit pipelines typically suffer from three sources of error accumulation: 
(1) localization errors from the detection model, 
(2) information loss and ambiguity introduced when converting bounding boxes into the handcrafted structured prompts, 
and (3) secondary encoding errors when VLM reinterprets these prompts.
Since final captions are produced by the LLM conditioned on visual evidence, we eliminate the intermediate prompt construction and re-encoding stages, 
and instead directly integrate the discriminative grounding features into LLM.
This design preserves efficiency while substantially reducing unnecessary transformation-induced degradation, and further enables the end-to-end optimization.

Peculiarly, as discussed in Section \ref{sec1}, 
existing feature adapters commonly focus on global embedding alignment, 
which is insufficient for grounding-based perception where detectors produce multi-source localization representations.
We therefore propose the Spatially-aware Grounding Adapter (SGA) to adaptively compress and aggregate these signals.

\begin{figure}[t]
    \centering
    \includegraphics[width=\linewidth]{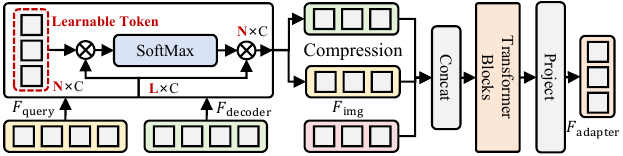}
    \caption{Detailed feature process pipeline of our designed SGA.
    }
    \label{adapter}
\end{figure}

\subsubsection{Adapter Inputs.}
As shown in Figure \ref{adapter}, 
our SGA receives three groups of grounding features as input: 
the original image features $F_{\text{img}}$, the text-guided cross-modal query features $F_{\text{query}}$, and the decoded discriminative grounding features $F_{\text{decoder}}$. 
These representations naturally capture complementary information, including global scene context, language-conditioned alignment, and object-centric discrimination,
which together provide rich and structured semantic cues for generating captions.

\subsubsection{Token Compression.}
Since $F_{\text{query}}, F_{\text{decoder}} \in \mathbb{R}^{L \times C}$ are produced by text-guided cross-modal querying and decoding, 
they represent the responses over all potential interest regions.
As a result, these features inevitably contain substantial redundancy and noise (e.g., background activations or false-positive regions). 
We therefore first perform token-level compression to distill discriminative-relevant semantics.
Specifically, we initialize a learnable token set $Q \in \mathbb{R}^{N \times C} (N \ll L)$ with the same dimension as $F_{\text{img}}$. 
The compressed representations are obtained via global attention modeling:
\begin{align}
\tilde{F}_{\text{query}} &= \text{SoftMax}(Q_{\text{query}} F_{\text{query}}^{\top})F_{\text{query}} \\
\tilde{F}_{\text{decoder}} &= \text{SoftMax}(Q_{\text{decoder}}F_{\text{decoder}}^{\top})F_{\text{decoder}}
\end{align}
Through end-to-end optimization, these learnable tokens can adaptively aggregate informative semantics 
while suppressing irrelevant activations, and meanwhile help effectively reduce the computational cost.

\subsubsection{Token Integration.}
After compression, we integrate these three heterogeneous discriminative features by concatenating them along the channel dimension:
\begin{equation} 
F_{\text{cat}} = \text{Concat}(F_{\text{img}}, \tilde{F}_{\text{query}}, \tilde{F}_{\text{decoder}})
\end{equation}
where global appearance cues, language-conditioned responses, and discriminative grounding information are explicitly aggregated into a unified token sequence. 
This concatenation preserves abundant semantic information and is then fed into the stacked Transformer blocks to perform deep semantic fusion and contextual refinement:
\begin{equation}
F_{\text{fuse}} = \text{Transformer}(F_{\text{cat}})
\end{equation}
This design allows cross-source interactions and self-attention-based reasoning among different visual and modal cues, 
yielding compact and context-aware representations.
Finally, a lightweight linear projection is applied to project the fused feature into the embedding space of the LLM:
\begin{equation}
F_{\text{adapter}} = \text{Linear}(F_{\text{fuse}})
\end{equation}
so that the adapted tokens can be seamlessly injected into the LLM as visual conditioning tokens, 
directly supporting high-level caption reasoning without introducing additional prompt engineering or re-encoding stages.

\begin{table*}[t]
\centering
\caption{Quantitative performance comparison between existing mainstream strategies and our GovLA-Reasoner on GovLA-10K.}
\setlength{\tabcolsep}{11pt}
\renewcommand{\arraystretch}{1.25}
\resizebox{\linewidth}{!}{
\begin{tabular}{l|ccccccc}
\hline \hline
Method & BLEU-1 & BLEU-2 & BLEU-3 & BLEU-4 & METEOR & ROUGE-L & CIDEr-D \\
\hline \hline
\multicolumn{8}{c}{MMGroundingDINO\cite{zhao2024open}+VLM+Explicit Prompt Construction} \\
\hline
LLaVA-OneVision-1.5-4B \cite{LLaVA-OneVision-1.5}   
& 36.27 & 21.24 & 12.56 & 7.61 & 19.10 & 25.36 & 4.84  \\

LLaVA-OneVision-1.5-8B \cite{LLaVA-OneVision-1.5}   
& 30.61 & 18.28 & 10.86 & 6.78 & 17.25 & 24.82 & 2.69  \\

InternVL3-8B \cite{zhu2025internvl3}          
& 31.72 & 17.27 & 9.14 & 5.14 & 17.68 & 22.39 & 2.72  \\

InternVL3.5-4B  \cite{wang2025internvl3_5}         
& 31.01 & 17.01 & 9.34 & 5.28 & 17.33 & 22.61 & 2.71  \\

InternVL3.5-8B \cite{wang2025internvl3_5}         
& 34.56 & 18.82 & 10.06 & 5.64 & 18.28 & 22.44 & 3.01  \\

Qwen2.5-VL-3B \cite{Qwen2.5-VL}             
& 37.17 & 21.25 & 12.71 & 8.04 & 19.21 & 25.01 & 5.26  \\

Qwen2.5-VL-7B \cite{Qwen2.5-VL}            
& 36.15 & 21.51 & 13.20 & 8.65 & 19.54 & 25.63 & 5.07  \\

Qwen3-VL-4B \cite{Qwen3-VL}               
& 45.77 & 27.72 & 17.41 & 11.36 & 23.25 & 28.92 & 10.22  \\

Qwen3-VL-8B  \cite{Qwen3-VL}             
& 40.88 & 25.64 & 16.54 & 10.97 & 21.73 & 29.44 & 10.21  \\
\hline \hline
\multicolumn{8}{c}{MMGroundingDINO\cite{zhao2024open}+LLM+Implicit Feature  Coordination} \\
\hline
Qwen3-4B\cite{qwen3}+LLaVA Projector\cite{liu2023visual}              
&48.96  &32.18  &22.02  &15.54  &24.95 &34.02  &10.56   \\
Qwen3-4B\cite{qwen3}+Q-Former\cite{li2023blip}        
&52.35  &36.50  &26.54  &19.94  &26.23  &37.72  & 19.62  \\
\hline \hline
\rowcolor{gray!15} GovLA-Reasoner (ours)  
& \textbf{53.32} & \textbf{37.10} & \textbf{26.98} & \textbf{20.31} & \textbf{26.63} & \textbf{37.97} & \textbf{20.31}  \\
\hline \hline
\end{tabular}}
\label{tab_cap}
\end{table*}
\begin{table}[t]
\centering
\setlength{\tabcolsep}{10pt}
\caption{Detection performance evaluation of different representative methods on GovLA-10K dataset.
}
\renewcommand{\arraystretch}{1.2}
\resizebox{\linewidth}{!}{
\begin{tabular}{l|ccc}
\hline \hline
Method & mAP & mAP@50 & mAP@75 \\
\hline
Faster R-CNN \cite{ren2016faster}        & 35.1 & 71.1 & 29.4 \\
Cascade R-CNN \cite{cai2018cascade}     & 37.2 & 71.2 & 34.7 \\
YOLO-X \cite{ge2021yolox}            & 31.6 & 65.3 & 25.2 \\
Deformable DETR \cite{zhu2020deformable}  & 36.6 & 69.3 & 34.3 \\
DINO \cite{zhang2022dino}               & 38.6 & 70.6 & 37.5 \\
\rowcolor{gray!15} MMGroundingDINO \cite{zhao2024open}    & \textbf{42.0} & \textbf{76.6} & \textbf{42.3} \\
\hline \hline
\end{tabular}}
\label{tab_det}
\end{table}

\begin{figure*}[!t]
    \centering
    \includegraphics[width=\linewidth]{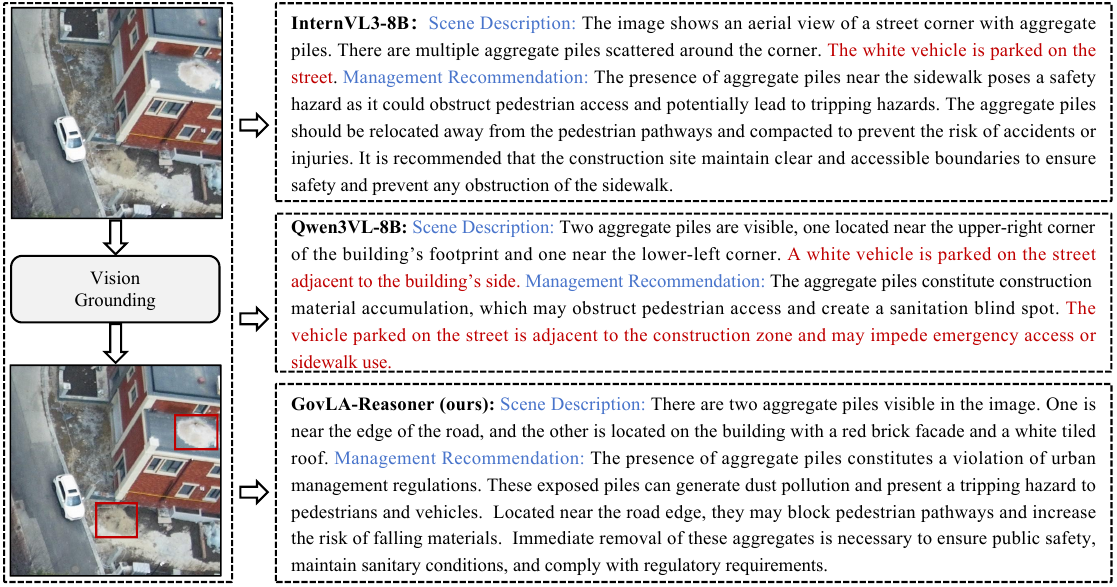}
\caption{Caption reasoning result comparison between existing mainstream VLMs and our GovLA-Reasoner.
   Unexpected contents are highlighted in red for clarity.}
    \label{pre}
\end{figure*}

\begin{table}[t]
\centering
\caption{Model complexity comparison between existing mainstream adapters and our SGA.
MSI represents multi-stream integration capability. 
}
\setlength{\tabcolsep}{3pt}
\renewcommand{\arraystretch}{1.2}
\resizebox{\linewidth}{!}{
\begin{tabular}{l|c|ccc}
\hline \hline
Adapter & MSI & Params(M) & FLOPs(G) & Peak Memory(MB)\\
\hline
LLaVA Projector\cite{liu2023visual} &\xmark &4.462   &9.127   &244.020  \\
Q-Former\cite{li2023blip} &\xmark &21.257   &43.520  & 382.637  \\
\hline
\rowcolor{gray!15} SGA (ours) & \cmark &\textbf{17.582}   &\textbf{36.002}   &\textbf{317.893}   \\
\hline \hline
\end{tabular}}
\label{ada_table}
\end{table}
\begin{table}[t]
\centering
\caption{Ablation study on the depth of SGA.}
\setlength{\tabcolsep}{10pt}
\renewcommand{\arraystretch}{1.2}
\resizebox{\linewidth}{!}{
\begin{tabular}{c|cccc}
\hline \hline
Depth & BLEU-4 & METEOR & ROUGE-L & CIDEr-D \\
\hline
1 & 19.54 & 26.29 & 37.56 & 18.68\\
3 & 20.28  &\textbf{26.71}  &37.68  &18.10  \\
4 & 19.94 &26.25  &37.90  &18.20  \\
\rowcolor{gray!15} 2 
& \textbf{20.31} & 26.63 & \textbf{37.97} & \textbf{20.31} \\
\hline \hline
\end{tabular}}
\label{ab2}
\end{table}

\begin{table}[t]
\centering
\caption{Ablation study on the contribution of different discriminative features.}
\setlength{\tabcolsep}{4pt}
\renewcommand{\arraystretch}{1.2}
\resizebox{\linewidth}{!}{
\begin{tabular}{ccc|cccc}
\hline \hline
$F_{img}$ & $F_{query}$ &$F_{decoder}$ & BLEU-4 & METEOR & ROUGE-L & CIDEr-D \\
\hline
\xmark &\cmark &\cmark & 17.97 & 25.00 & 36.17 & 14.54 \\
\cmark &\xmark &\cmark & 19.62 & 26.31 & 37.51 & 17.81 \\
\cmark &\cmark &\xmark & 19.81 & 26.46 & 37.45 & 18.36 \\
\rowcolor{gray!15} \cmark & \cmark  & \cmark
& \textbf{20.31} & \textbf{26.63} & \textbf{37.97} & \textbf{20.31} \\
\hline \hline
\end{tabular}}
\label{ab3}
\end{table}
\subsubsection{Adapter Optimization.}  
Considering the core purpose of SGA is to transform the grounding-aware visual features into LLM-compatible representations, 
it is sufficient to optimize SGA solely using the cross-entropy loss between generated captions and ground-truth captions. 
Formally, the training objective can be written as:
\begin{equation}
\mathcal{L}_{\text{CE}} = - \frac{1}{T} \sum_{t=1}^{T} \log P(y_t \mid y_{<t}, F_{\text{adapter}})
\end{equation}
where $y_t$ denotes the $t$-th token in the ground-truth caption, $y_{<t}$ represents the preceding tokens, and $F_{\text{adapter}}$ denotes the grounding-aware visual features after adapter transformation.
Importantly, all other components, including the grounding model and the LLM itself, are kept frozen during training. 
This targeted optimization strategy effectively constrains SGA while avoiding perturbation of the pre-trained representations. 
Moreover, by limiting gradient updates to the lightweight adapter, we can achieve efficient training with minimal computational overhead, 
ensuring both practicality and scalability for large-scale deployment.


\section{Experiment}

Considering the absence of other publicly available datasets in low-altitude urban management, 
we perform experiments primarily on the proposed GovLA-10K dataset. 
To establish a comprehensive analysis, 
we first select several representative visual grounding algorithms to assess their performance. 
Based on the evaluation results, we choose the most advanced method as the baseline, which is then combined with the mainstream VLMs and LLM to perform the caption reasoning task.
Through this experimental pipeline, 
we aim to demonstrate both the applicability of GovLA-10K and the effectiveness of GovLA-Reasoner for low-altitude aerial perception and reasoning.

\subsection{Detail Settings}
\subsubsection{Implementation Details.}
All experiments are implemented under the PyTorch framework \cite{paszke2019pytorch} and conducted on eight NVIDIA H20 GPUs. 
We deploy Qwen3-4B \cite{qwen3} as the LLM component of GovLA-Reasoner.
We adopt the default loss configurations provided by the MMDetection \cite{chen2019mmdetection} and the Transformers libraries \cite{wolf2020transformers} to perform optimization.
The batch size is set to 16, and AdamW is used as the optimizer with an initial learning rate of $1 \times 10^{-4}$. 
The multi-step learning rate scheduler is employed to decay the learning rate at predefined epochs during training. 
For the detection task, the model is trained for 50 epochs, while for the captioning task, the max epoch is set to 25. 
Considering the availability of open-source implementations, 
three representative VLMs \cite{LLaVA-OneVision-1.5,zhu2025internvl3,wang2025internvl3_5,Qwen2.5-VL,Qwen3-VL}
and two representative adapters \cite{liu2023visual,li2023blip}
are selected for comparison, covering different versions and parameter scales.
All compared methods follow the same configuration to ensure fair comparisons.
\subsubsection{Evaluation Metrics.}
For the detection task, we report mean Average Precision (mAP) at different IoU thresholds, 
including mAP, mAP@50, and mAP@75, to comprehensively evaluate the model performance.
For the captioning task, we employ four widely used natural language generation metrics, 
including BLEU-N (N=1,2,3,4) \cite{papineni2002bleu}, ROUGE-L \cite{lin2004rouge}, METEOR \cite{banerjee2005meteor} and CIDEr-D \cite{vedantam2015cider}, to assess the quality of the generated descriptions from multiple perspectives. 
Higher values of all these metrics indicate better model performance.

\subsection{Main Results}

\subsubsection{Quantitative Results.}

Table~\ref{tab_det} reports the quantitative results of several representative detection frameworks on the proposed GovLA-10K dataset. 
Instead of focusing on architectural modifications, we deliberately select a set of widely adopted and competitive detectors, 
including two-stage, one-stage, and transformer-based methods, to provide a comprehensive and objective evaluation. 
As shown in the table, most existing detectors achieve reasonably strong performance, indicating that current detection methods are already sufficient for handling low-altitude urban scenarios.
This observation is consistent with our initial motivation that the bottleneck of low-altitude understanding does not mainly lie in pure object detection. 
Notably, MMGroundingDINO achieves the best results across all metrics. 
We attribute this advantage to its ability to incorporate textual prompts as inputs, 
which enables more selective and goal-driven localization. 
This property is highly aligned with the intrinsic characteristics of GovLA-10K, where interest objects are often defined by management-oriented semantics rather than purely visual categories. 

Table~\ref{tab_cap}  presents the quantitative captioning metrics of existing mainstream strategies and our proposed GovLA-Reasoner on GovLA-10K dataset. 
Because MMGroundingDINO in Table \ref{tab_det} achieves the best detection performance, 
we thus deploy it to support the caption reasoning, which can provide a stable grounding foundation for the subsequent captioning reasoning. 
A clear observation is that the VLM scale is not consistently correlated with performance. 
Several larger models (e.g., 8B variants of LLaVA-OneVision-1.5 and Qwen3-VL) do not outperform, and in some cases underperform, their smaller counterparts. 
This is because the existing loosely coupled paradigm inevitably discards fine-grained spatial structure and amplifies upstream localization noise during secondary encoding.
In contrast, introducing feature adapters brings clear improvements. 
For instance, combining MMGroundingDINO with Qwen3-4B through Q-Former significantly outperforms 
explicit prompt-based pipelines, 
indicating that implicit feature-level coordination between grounding representations 
and LLM is beneficial.

However, compared with all the methods in the table, 
GovLA-Reasoner achieves substantial and consistent improvements across all evaluation metrics. 
This is because our SGA can effectively integrate multi-stream discriminative grounding features into LLM, 
allowing continuous spatial cues, language-conditioned responses, and object-centric representations to be jointly leveraged. 
This design enables LLM to exploit visual evidence in a more faithful and coherent manner, 
resulting in more accurate, detailed, and management-oriented descriptions. 
Peculiarly, although GovLA-Reasoner only employs a 4B-scale LLM, it still significantly outperforms all compared VLMs, including multiple 8B models. 
This demonstrates that the proposed framework not only improves effectiveness, but also achieves superior parameter efficiency, highlighting the advantage of our design.

Table~\ref{ada_table} further reports the model complexity of our proposed SGA. 
Although SGA introduces the ability to integrate multi-stream grounding features, 
its computational overhead remains moderate. 
Specifically, SGA requires only 17.582M parameters and 36.002G FLOPs, 
which are substantially lower than those of Q-Former while providing stronger representational capability. 
Meanwhile, the peak GPU memory consumption remains well controlled (317.893 MB), 
indicating that our adapter introduces limited additional computational burden. 
This is because SGA introduces the efficient token compression mechanism to distill discriminative semantics.
Therefore, these quantitative results once again confirm the efficiency and practicality of our design.

\subsubsection{Qualitative Results.}
Figure \ref{pre} presents the qualitative comparison between several mainstream VLMs and our proposed GovLA-Reasoner on one low-altitude urban scene. 
Overall, existing VLMs are able to generate semantically rich and fine-grained captions, 
demonstrating their strong generalization ability and powerful open-world visual understanding. 
As highlighted by the red text in the figure, 
the core management concern in this scenario is the identification and analysis of aggregate piles.
However, existing VLMs commonly tend to produce over-generalized or misaligned reasoning results, 
such as shifting attention to irrelevant objects (e.g., vehicles), or drawing conclusions unrelated to the actual risk source. 
In contrast, our GovLA-Reasoner always produces more precise, goal-consistent, and practically meaningful descriptions. 
Therefore, this qualitative comparison once again intuitively verifies the effectiveness of the proposed framework and its advantages in real-world low-altitude scene understanding.

\subsection{Ablation Study}
Our ablation experiments primarily focus on the design details of the proposed SGA.
The specific experimental results are presented below.

\subsubsection{Depth of Adapter.}
As shown in Table~\ref{ab2}, 
because we deploy stacked Transformer blocks to perform the feature refinement,
the block number directly determines how thoroughly heterogeneous grounding features can be aligned and fused.
Obviously, using a single layer yields limited gains due to insufficient cross-source interaction, 
while excessively deep adapters bring marginal or even negative effects. 
The best overall performance is achieved with two layers, suggesting that moderate semantic fusion is sufficient.

\subsubsection{Contribution of Discriminative Features.}
Because our adapter takes three groups of grounding-aware features as input, 
we remove each component individually to evaluate its contribution. 
As shown in Table~\ref{ab3}, excluding any feature consistently degrades performance, 
confirming the rationality of our multi-source design.
Notably, removing $F_{\text{img}}$ causes the most significant performance drop.
This is because the original image feature preserves the most comprehensive and fundamental visual information, 
providing holistic scene context that cannot be fully recovered from the derived discriminative representations alone. 

\section{Conclusion}
This work aims to address the critical gaps in management-oriented low-altitude intelligence.
On the one hand, we introduced GovLA-10K, 
the first multimodal benchmark designed around functionally salient, governance-relevant targets, shifting low-altitude research from generic perception to selective and semantics-driven reasoning.
On the other hand, we propose GovLA-Reasoner, a unified vision–language reasoning framework that implicitly coordinates visual grounding and language generation via the tailored grounding adapter. 
By directly projecting grounding-aware visual representations into the LLM feature space, 
GovLA-Reasoner avoids the limitations of the explicit prompt construction pipeline.
Extensive experiments demonstrate that GovLA-Reasoner consistently outperforms strong mainstream VLM baselines across all captioning metrics, highlighting both its effectiveness and deployment efficiency. 
We believe this work establishes a new foundation for management-aware low-altitude vision–language systems and opens promising directions for future research in governance-oriented aerial intelligence.

\bibliographystyle{IEEEtran}
\bibliography{sample-base}

\end{document}